\documentclass[letterpaper, 10 pt, conference]{ieeeconf}  

\IEEEoverridecommandlockouts                              

\usepackage{graphics} 
\usepackage{epsfig} 
\usepackage{caption}
\usepackage{subcaption}
\graphicspath{ {figs} }
\usepackage{times} 
\usepackage{amsmath} 
\usepackage{amssymb}  
\usepackage{textcomp}
\usepackage{color}
\usepackage{soul}
\usepackage{nth}
\usepackage{multirow}
\usepackage{siunitx}
\newcommand{\etal}{{\em et al. }}

\newcommand{\ba}{\mathbf{a}}
\newcommand{\bs}{\mathbf{s}}

\newcommand{\norm}[1]{\left\lVert#1\right\rVert}
\DeclareMathOperator{\atantwo}{atan2}

\title{\LARGE \bf
Curiosity-driven Exploration for Mapless Navigation
\\with Deep Reinforcement Learning
}

\author{Oleksii Zhelo$^1$, \
Jingwei Zhang$^1$, \
Lei Tai$^2$, \
Ming Liu$^2$, \
Wolfram Burgard$^1$ 
\thanks{$^{1}$Department of Computer Science, Albert Ludwig University of Freiburg.
        {\tt\small oleksii.zhelo@saturn.uni-freiburg.de, \{zhang, burgard\}@informatik.uni-freiburg.de}}%
\thanks{$^{2}$Department of Electronic and Computer Engineering, The Hong Kong University of Science and Technology.
        {\tt\small \{ltai, eelium\}@ust.hk}}%
}

\begin{document}
\maketitle
\thispagestyle{empty}
\pagestyle{empty}

\begin{abstract}

This paper investigates exploration strategies of Deep Reinforcement Learning (DRL) methods to learn navigation policies for mobile robots.
In particular, we augment the normal external reward for training DRL algorithms with intrinsic reward signals measured by curiosity.
We test our approach in a mapless navigation setting,
where the autonomous agent is required to navigate without the occupancy map of the environment,
to targets whose relative locations can be easily acquired through low-cost solutions
(e.g., visible light localization, Wi-Fi signal localization).
We validate that the intrinsic motivation
is crucial for improving DRL performance in tasks with challenging exploration requirements.
Our experimental results show that our proposed method is able to more effectively learn navigation policies,
and has better generalization capabilities in previously unseen environments.
A video of our experimental results can be found at https://goo.gl/pWbpcF.



\end{abstract}

\section{Introduction}
\label{sec:introduction}
Deep Reinforcement Learning (DRL),
deploying deep neural networks as function approximators for high-dimensional RL tasks,
achieves state of the art performance in various fields of research \cite{mnih2016asynchronous}.

\begin{figure}[t]
    \centering
    \includegraphics[width=0.8\columnwidth]{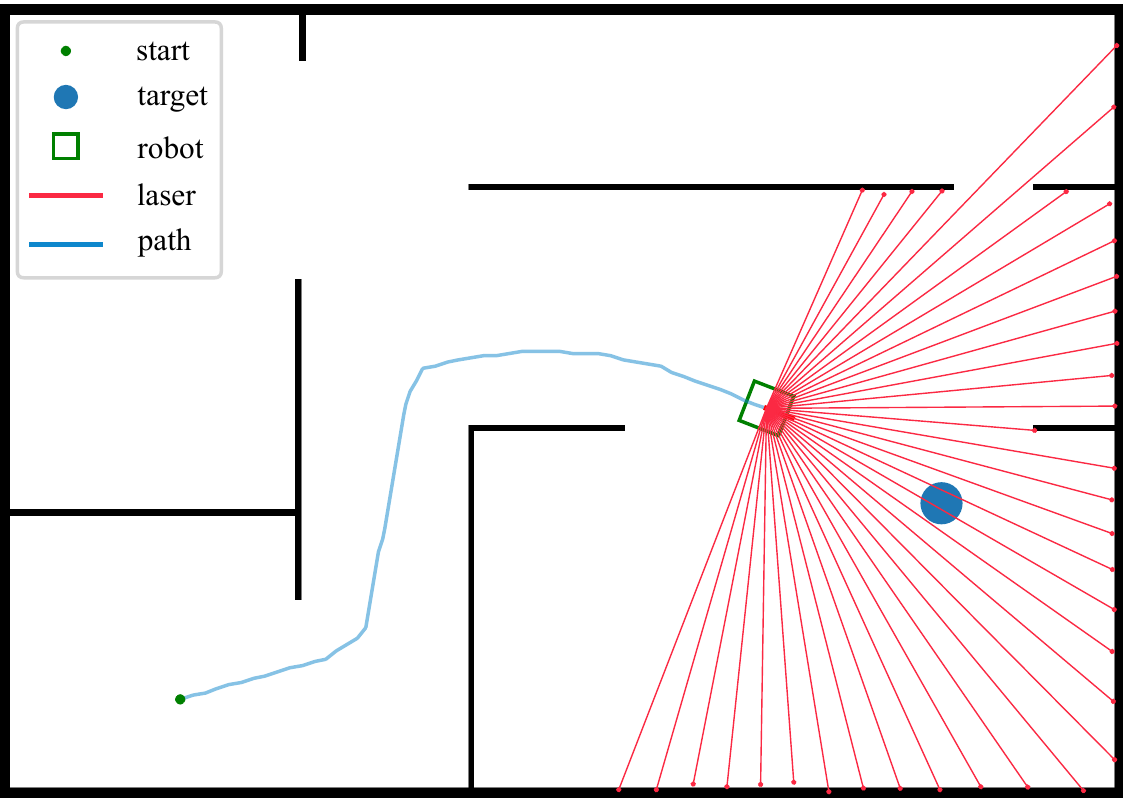}
    \caption{We study the problem of mapless navigation,
    where without a given the map of the environment,
    the autonomous agent is required to navigate to a target
    whose relative location can be easily acquired by cheap localization solutions.
    Structures like long corridors and dead corners make it challenging
    for DRL agents to learn optimal navigation policies.
    Trained with reward signals that are augmented by intrinsic motivation,
    our proposed agent efficiently tackles this task,
    and shows that its learned policy generalizes well to previously unseen environments.}
    \label{fig:title}
\end{figure}

DRL algorithms have been studied under the context of learning navigation policies for mobile robots.
Traditional navigation solutions in robotics generally require a system of procedures,
such as Simultaneous Localization and Mapping (SLAM) \cite{thrun2005probabilistic},
localization and path planning in a given map, etc.
With the powerful representation learning capabilities of deep networks,
DRL methods bring about the possibility of learning control policies directly from raw sensory inputs,
bypassing all the intermediate steps.


Eliminating the requirement for localization, mapping, or path planning procedures,
several DRL works have been presented that learn successful navigation policies directly from raw sensor inputs:
target-driven navigation \cite{zhu2017target},
successor feature RL for transferring navigation policies \cite{zhang2017deep},
and using auxiliary tasks to boost DRL training \cite{mirowski2016learning}.
Many follow-up works have also been proposed,
such as embedding SLAM-like structures into DRL networks \cite{zhang2017neural},
or utilizing DRL for multi-robot collision avoidance \cite{long2017towards}.

In this paper,
we focus specifically on mapless navigation,
where the agent is expected to navigate to a designated goal location
without the knowledge of the map of its current environment.
We assume that the relative pose of the target is easily acquirable for the agent
via cheap localization solutions
such as visible light localization \cite{liang2017plugo} or Wi-Fi signal localization \cite{sun2014wifi}.
Tai \etal \cite{tai2017virtual} successfully applied DRL for mapless navigation,
taking as input sparse laser range readings,
as well as the velocity of the robot and the relative target position.
Trained with asynchronous DRL,
the policy network outputs continuous control commands for a nonholonomic mobile robot,
and is directly deployable in real-world indoor environments.

Most of the aforementioned methods, however,
either rely on random exploration strategies like $\epsilon$-greedy,
or on state-independent exploration by maximizing the entropy of the policy.
As the previous works present experiments in environments which do not impose considerable challenges for the exploration of DRL algorithms, we suspect that these exploration approaches might not be sufficient for learning efficient navigation policies in more complex environments.

Proposed by Pathak \etal \cite{pathak2017curiosity},
the agent's ability to predict the consequences of its own actions
can be used to measure the novelty of the states, or the intrinsic curiosity.
The \textit{Intrinsic Curiosity Module} (ICM) is employed to acquire this signal during the process of reinforcement learning,
and its prediction error can serve as an intrinsic reward.
As illustrated in their experiments, the intrinsic reward, learned completely in a self-supervised manner,
can motivate the agent to better explore the current environment,
and make use of the structures of the environment for more efficient planning strategies.

Similar intrinsic signals have also been used as auxiliary tasks to encourage exploration by Khan \etal \cite{2018arXiv180301846K}, where a planning agent is developed using differentiable memory and self-supervised state, reward and action prediction. 

In this paper,
we investigate exploration mechanisms for aiding DRL agents
to learn successful navigation policies in challenging environments
that might contain structures like long corridors and dead corners.
We conduct a series of experiments in simulated environments,
and show the sample-efficiency,
stability,
and generalization ability of our proposed agent.


\section{Methods}
\label{sec:methods}

\subsection{Background}
We formulate the problem of autonomous navigation as a \textit{Markov Decision Process} (MDP),
where at each time step $t$, the agent receives an observation
of its current state $\bs_t$,
takes an action $\ba_t$, receives a reward $R_t$,
and transits to the next state $\bs_{t+1}$ following the transition dynamics $p(\bs_{t+1}|\bs_t,\ba_t)$ of the environment.
For the mapless navigation task that we consider,
the state $\bs_t$ consists of laser range readings $\bs^l_t$ and the relative pose of the goal $\bs^g_t$.
The task of the agent is to reach the goal position $g$ without colliding with obstacles.

\subsection{Extrinsic Reward for Mapless Navigation}
\label{sssec:external}

We define the extrinsic reward $R^{e}$ (the traditional reinforcement signal received from the environment) at timestep t as follows ($\lambda^p$ and $\lambda^{\omega}$ denote the scaling factors):
\begin{align}
	R^{e}_{t} = r_t + \lambda^{p} r^{p}_t + \lambda^{\omega} r^{\omega}_t,
\end{align}
where $r_t$ imposes the main task ($\mathbf{p}_t$ represents the pose of the agent at $t$):
\[
r_t =
\begin{cases}
r_{\text{reach}},& \text{if reaches goal,} \\
r_{\text{collision}}, & \text{if collides,} \\
\lambda^{g} \left(\norm{\mathbf{p}^{x,y}_{t-1} - g}_2 - \norm{\mathbf{p}^{x,y}_{t} - g}_2\right),  & \text{otherwise,}
\end{cases}
\]
and the position- and orientation-based penalties $r^{p}_t$ and $r^{\omega}_t$ are defined as:
\begin{align}
    r^{p}_t
&=
    \begin{cases}
    r_{\text{position}},& \text{if } \norm{\mathbf{p}^{x,y}_{t-1} - \mathbf{p}^{x,y}_{t}}_2 = 0, \\
    0, & \text{otherwise,}
    \end{cases}
\nonumber\\
    r^{\omega}_t
&=
    \norm{ \atantwo(\mathbf{p}_{t}^y - g^y, \mathbf{p}_{t}^x - g^x) - \mathbf{p}^{\omega}_{t} }_1.
\nonumber
\end{align}

\subsection{Intrinsic Reward for Curiosity-driven Exploration}
\label{sssec:intrisic}
On top of the normal extrinsic reward $R^{e}$,
we use intrinsic motivation measured by curiosity, rewarding novel states to encourage exploration
and consequently improve the learning and generalization performance of DRL agents in environments that require more guided exploration.

\begin{figure}[t]
	\centering
	\includegraphics[width=0.9\columnwidth]{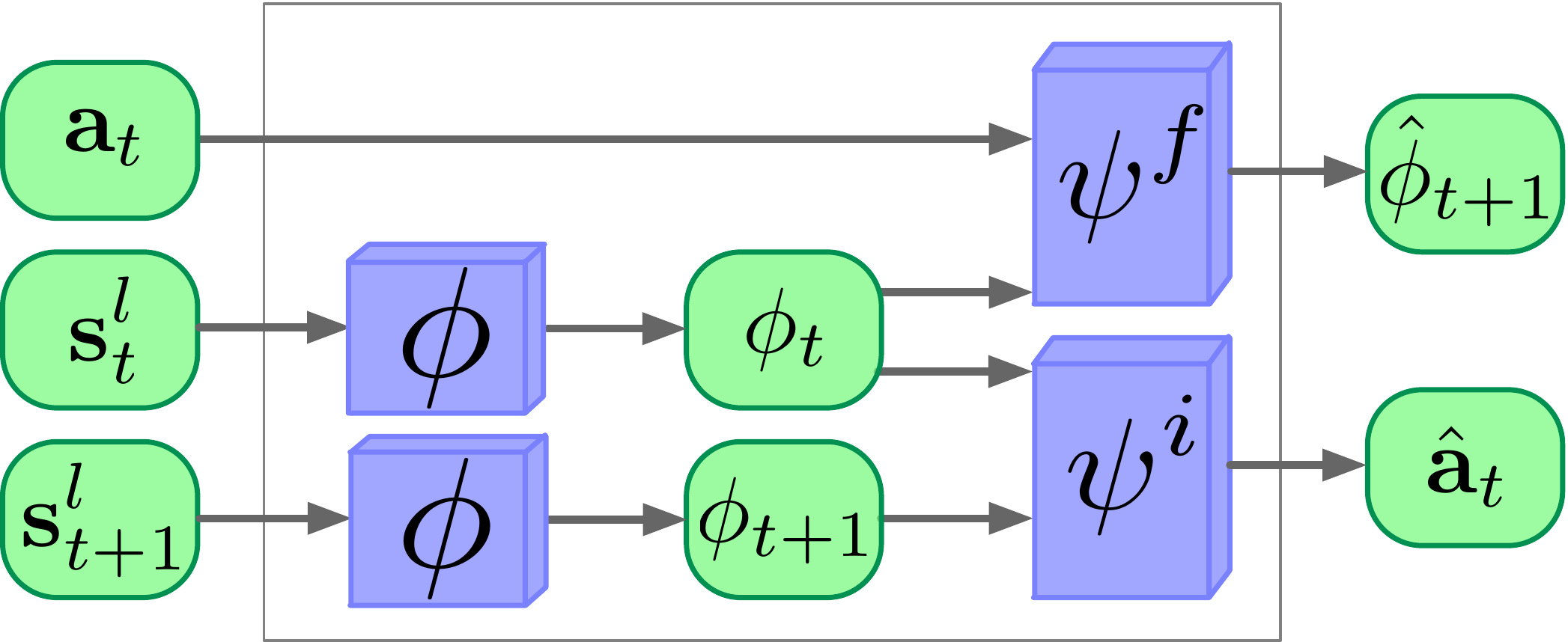}
	\caption{ICM architecture.
		$\bs^l_t$ and $\bs^l_{t+1}$ are first passed through the feature extraction layers $\phi$,
		and encoded into $\phi_t$ and $\phi_{t+1}$.
		Then $\phi_t$ and $\phi_{t+1}$ are input together into the \textit{inverse model} $\psi^i$,
		to infer the action $\hat{\ba}_t$.
		At the same time, $\ba_t$ and $\phi_t$ are together used to predict $\hat{\phi}_{t+1}$,
		through the \textit{forward model} $\psi^f$.
		The prediction error between $\hat{\phi}_{t+1}$ and ${\phi}_{t+1}$ is used as the intrinsic reward $R^i$.
	}
	\label{fig:net}
\end{figure}

Following the formulation of \cite{pathak2017curiosity},
we measure the intrinsic reward $R^{i}$ via an \textit{Intrinsic Curiosity Module} (ICM)
(depicted in Fig. \ref{fig:net}).
It contains several feature extraction layers $\phi$,
a \textit{forward model} (parameterized by $\psi^f$),
and an \textit{inverse model} (parameterized by $\psi^i$).

First, $\bs^l_t$ and $\bs^l_{t+1}$ are passed through $\phi$,
encoded into their corresponding features $\phi_t$ and $\phi_{t+1}$.
Then, $\psi^i$ predicts $\hat{\ba_t}$ from $\phi_t$ and $\phi_{t+1}$
(trained through cross-entropy loss for discrete actions with the ground truth ${\ba_t}$),
while $\psi^f$ predicts $\hat{\phi}_{t+1}$ from $\phi_t$ and $\ba_t$
(trained through mean squared error loss with the ground truth ${\phi}_{t+1}$).
$\phi$, $\psi^f$ and $\psi^i$ are learned and updated together with
the actor-critic parameters $\theta^{\pi}$ and $\theta^v$ (Sec. \ref{sec:a3c}) during the training process.

The overall optimization objective is  $$\min_{\theta_{\phi}, \theta_{\psi_f}, \theta_{\psi_i}}((1 - \lambda^f)(-\sum_{j=1}^{3} \ba_j \log\hat{\ba_j}) + \lambda^f \frac{1}{2} \left|\left| \hat{\phi}_{t+1} - \phi_{t+1} \right|\right|^2_2),$$
where the first part corresponds to the inverse model training objective, and the second part to the forward model prediction error. These losses are weighted by the $\lambda^f$ parameter, which helps control the trade-off between the characteristics both models are conferring to the feature extraction layers: 1) the inverse model seeks to extract information useful to predict the action taking the agent between consecutive states, and 2) the forward model strives to encourage extracting more predictable feature embeddings. To avoid learning constant features for all states (which would be trivially predictable), the magnitude of forward model's contribution to the learning process needs to be balanced by the weight $\lambda^f$. Both of these influences, however, help to exclude environmental factors which are unaffected by and do not affect the agent, so it would not stay constantly curious of irrelevant or inherently unpredictable details of the environment.

The prediction error between $\hat{\phi}_{t+1}$ and $\phi_{t+1}$ through
$\psi^f$ is then used as the intrinsic reward $R^i$:
\begin{align}
    R^i
&=
    \frac{1}{2} \left|\left| \hat{\phi}_{t+1} - \phi_{t+1} \right|\right|^2_2.
\end{align}

Following this intrinsic reward,
the agent is encouraged to visit novel states in the environment,
which is crucial in guiding the agent to get out of local minimums
or premature convergence of sub-optimal policies.
This exploration strategy allows the agent to make better use of the learned environment dynamics
to accomplish the task at hand.

\subsection{Asynchronous Deep Reinforcement Learning}
\label{sec:a3c}
For training the navigation policies, we follow the asynchronous advantage actor-critic (A3C) algorithm \cite{mnih2016asynchronous},
with the weighted sum of the external reward and the intrinsic reward as the supervision signal:
\begin{align}
    R=R^{e}+\lambda^i R^{i},
\end{align}
where $\lambda^i>0$ is a scaling coefficient.

A3C updates both the parameters of a policy network $\theta^{\pi}$ and a value network $\theta^{v}$,
to maximize the cumulative expected reward.
The value estimate is bootstrapped from $n$-step returns with a discount factor $\gamma$
(where $K$ represents the maximum number of steps for a single rollout):
\begin{align}
    G_t
&=
    \gamma^{K-t} V(\bs_{K};\theta^v) + \sum_{\tau=t}^{K-1}\gamma^{\tau-t}R_{\tau}.
\end{align}

Each learning thread accumulates gradients for every experience contained in a $K$ step rollout,
where the gradients are calculated according to the following equations:
\begin{align}
    d\theta^{\pi}
&=
    \nabla_{\theta^{\pi}} \log \pi(\ba_{t}|\bs_{t}; \theta^{\pi}) (G_{t} - V(\bs_t; \theta^{v}))
\\&\hspace{0.17in}+
    \beta \nabla_{\theta^{\pi}}H(\pi(\ba_{t}|\bs_{t};\theta^{\pi})),
\label{equ:policygrad}\\
    d\theta^{v}
&=
    \partial(G_{t} - V(\bs_{t};\theta^{v}))^{2} / \partial\theta^{v},
\label{equ:valuegrad}
\end{align}
where $H$ is the entropy of the policy,
and $\beta$ represents the coefficient of the entropy regularization.
This discourages premature convergence to suboptimal deterministic policies,
and is the default exploration strategy that has been used with A3C.
\section{Experiments}
\label{sec:experiments}

\subsection{Experimental setup}
\begin{figure}[t]
    \centering
        \begin{subfigure}{0.43\linewidth}
            \centering
            \includegraphics[width=\linewidth]{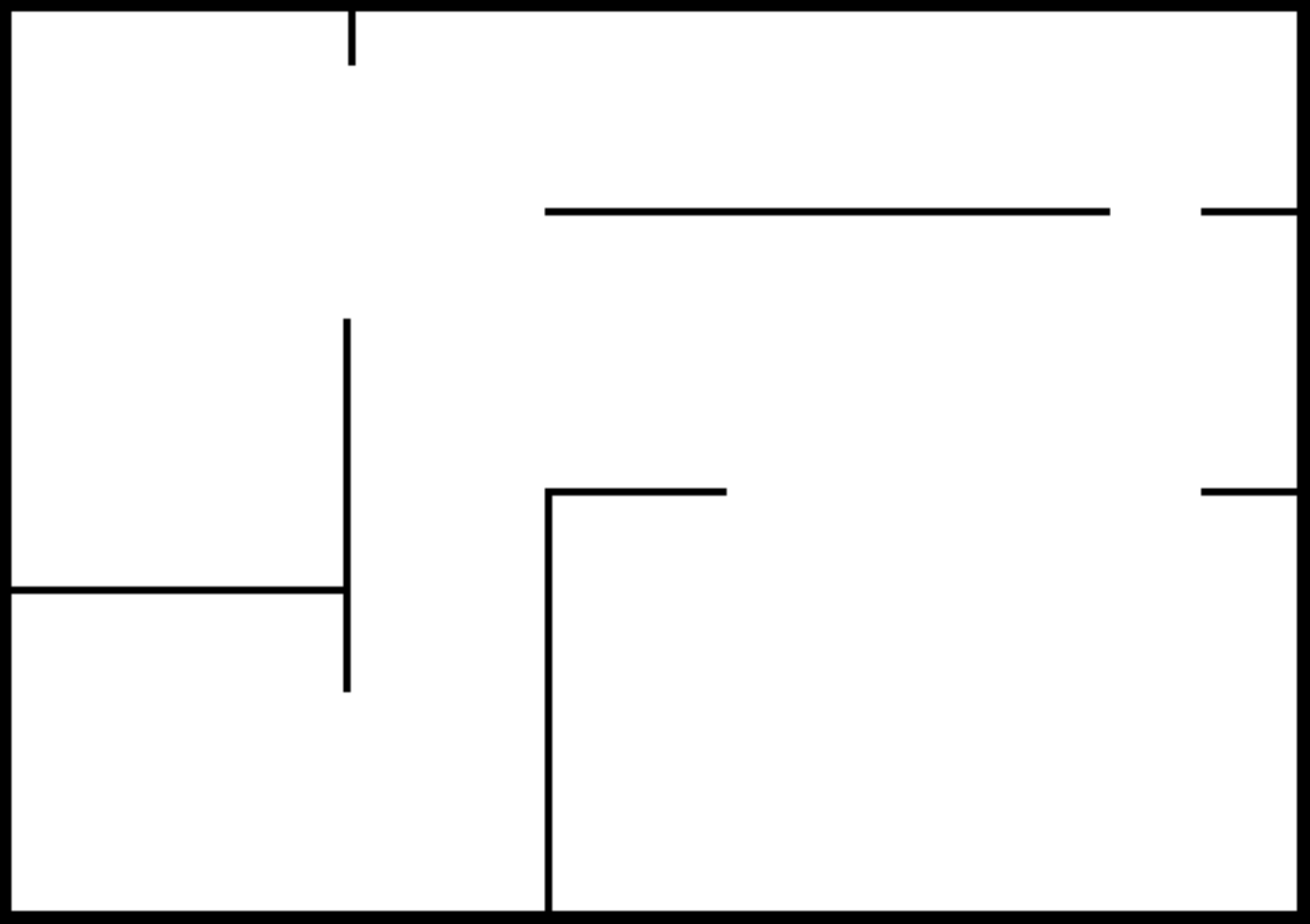}
            \caption{\textit{Map1}}
            \label{fig:map1}
        \end{subfigure}
        \begin{subfigure}{0.43\linewidth}
            \centering
            \includegraphics[width=\linewidth]{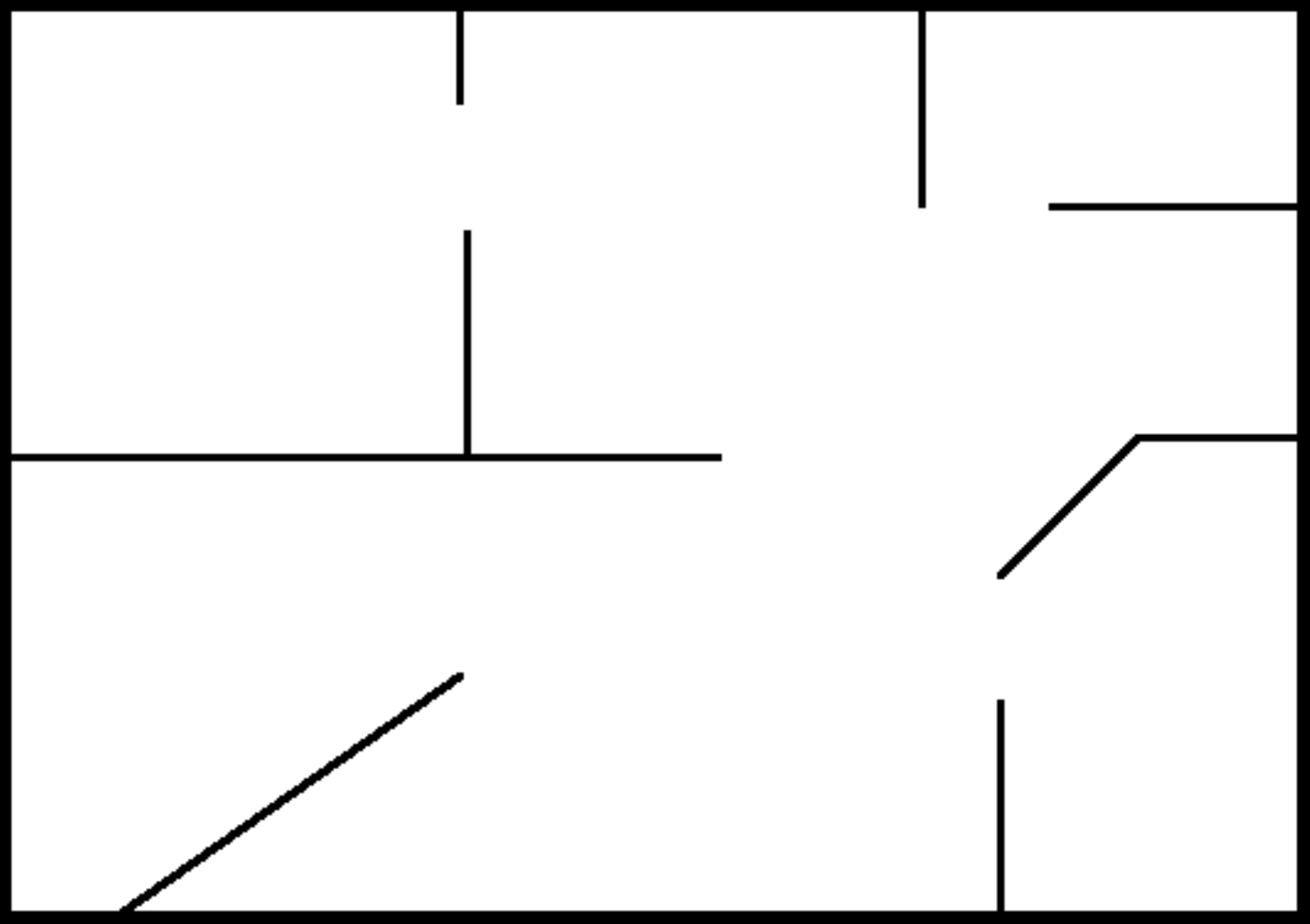}
            \caption{\textit{Map2}}
            \label{fig:map2}
        \end{subfigure} \\
        \begin{subfigure}{0.43\linewidth}
            \centering
            \includegraphics[width=\linewidth]{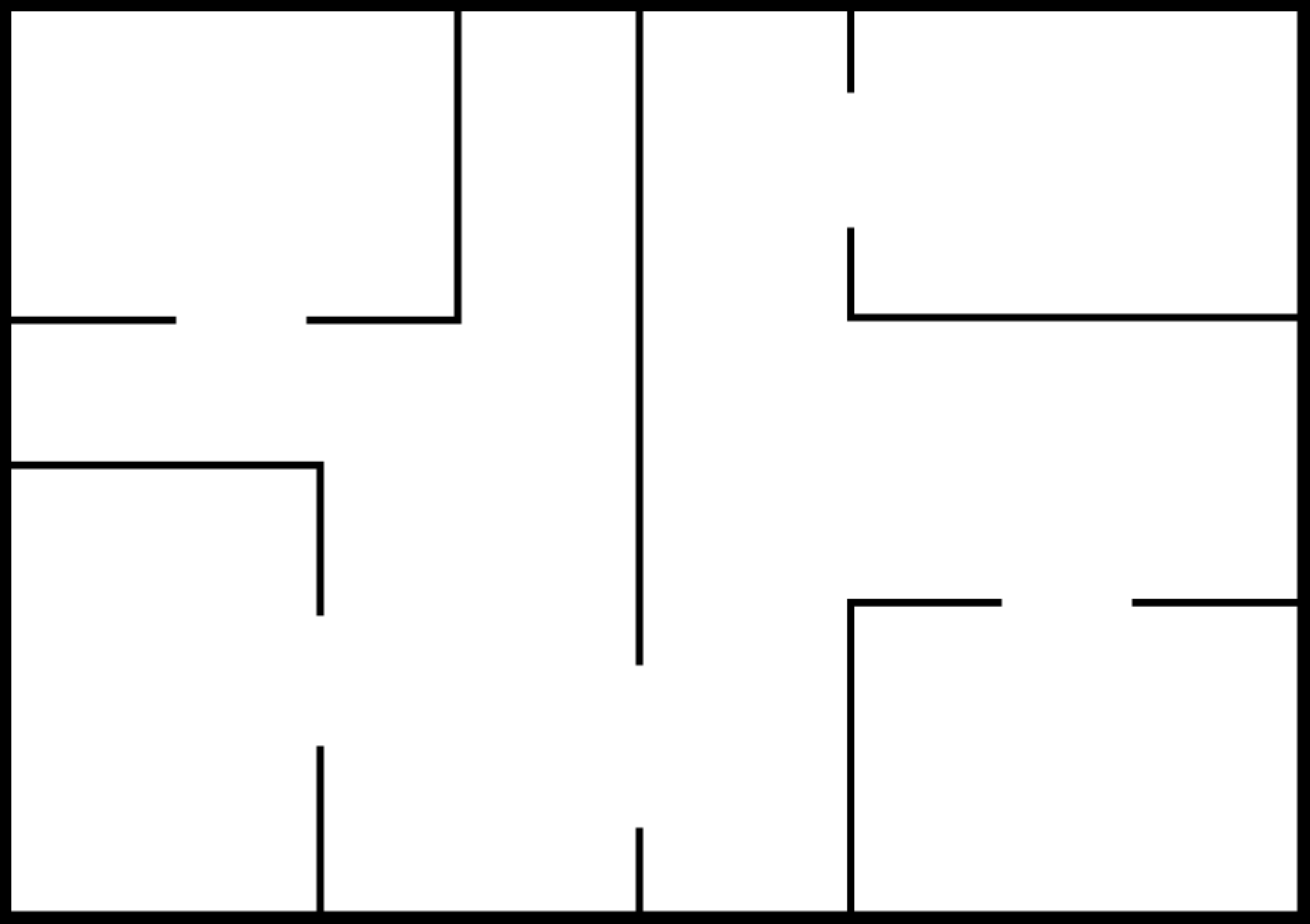}
            \caption{\textit{Map3}}
            \label{fig:map3}
        \end{subfigure}
        \begin{subfigure}{0.43\linewidth}
            \centering
            \includegraphics[width=\linewidth]{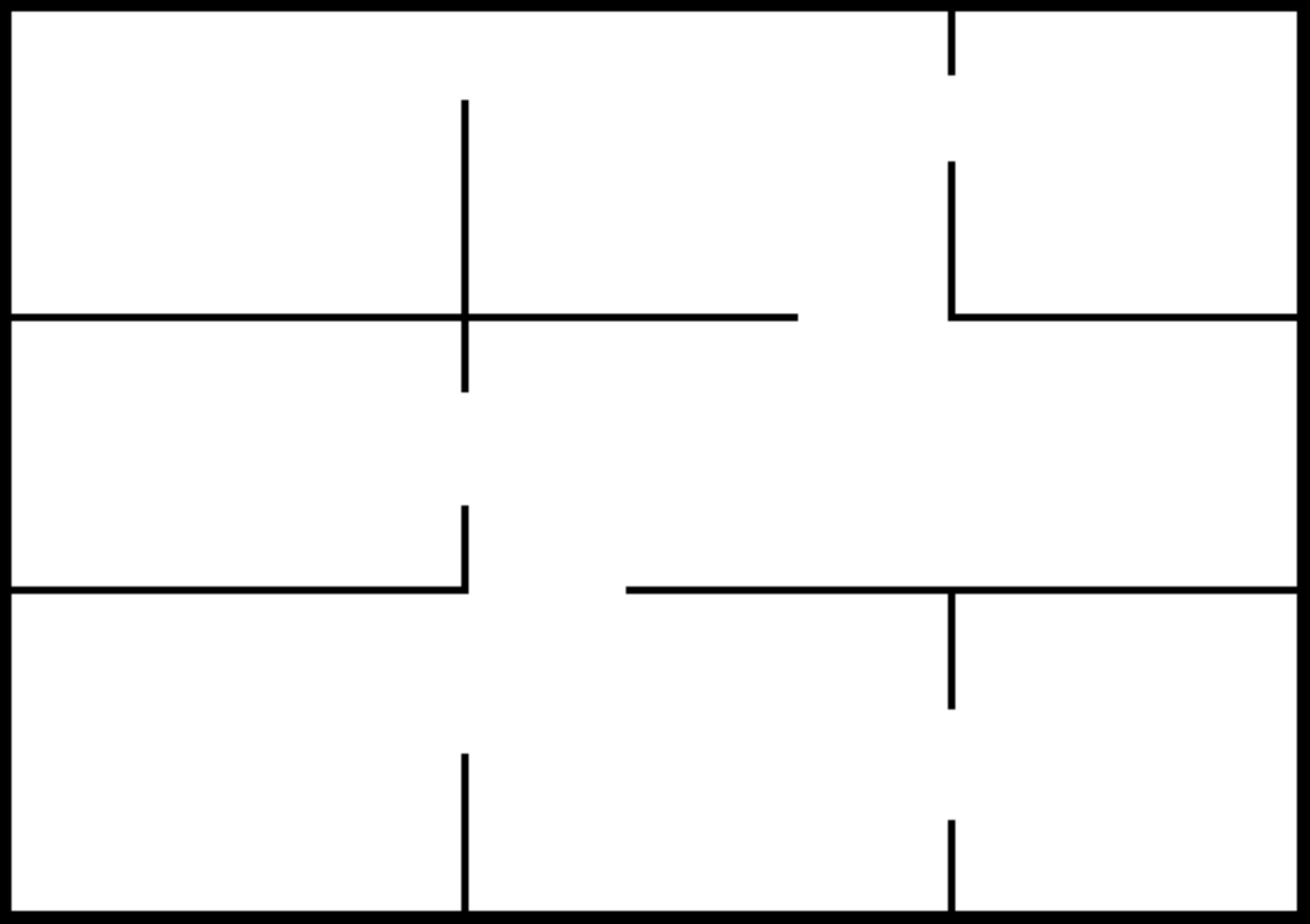}
            \caption{\textit{Map4}}
            \label{fig:map4}
        \end{subfigure}
    \caption{Different floor plans we considered in our experiments.
    \textit{Map1,3,4} have similar wall structures,
    while imposing gradually more exploration challenges for DRL agents to effectively learn navigation policies.
    \textit{Map2}, while holding a level of challenge for exploration similar to that of \textit{Map1},
    has some structural variations such as the angled walls.
    We only train agents on \textit{Map1}.
    The performance of the trained agents is later evaluated on \textit{Map1},
    and tested on \textit{Map2,3,4} for their generalization capabilities.
    The maps are all of size $\SI{5.33}{\metre}\times\SI{3.76}{\metre}$.}
    \label{fig:maps}
\end{figure}

We conduct our experiments in a simulated environment,
where a robot is equipped with a laser range sensor, and is
navigating in the $2D$ environments shown in Fig. \ref{fig:maps}.
At the beginning of each episode,
the starting pose of the agent $\mathbf{p}_0$
and the target position $g$ are randomly chosen such that
a collision-free path is guaranteed to exist between them.
An episode is terminated after the agent either reaches the goal, collides with an obstacle,
or after a maximum of $7000$ steps during training
and $400$ for testing.

\begin{figure}[t]
    \centering
        \begin{subfigure}[t]{0.9\linewidth}
            \centering
            \includegraphics[width=\linewidth]{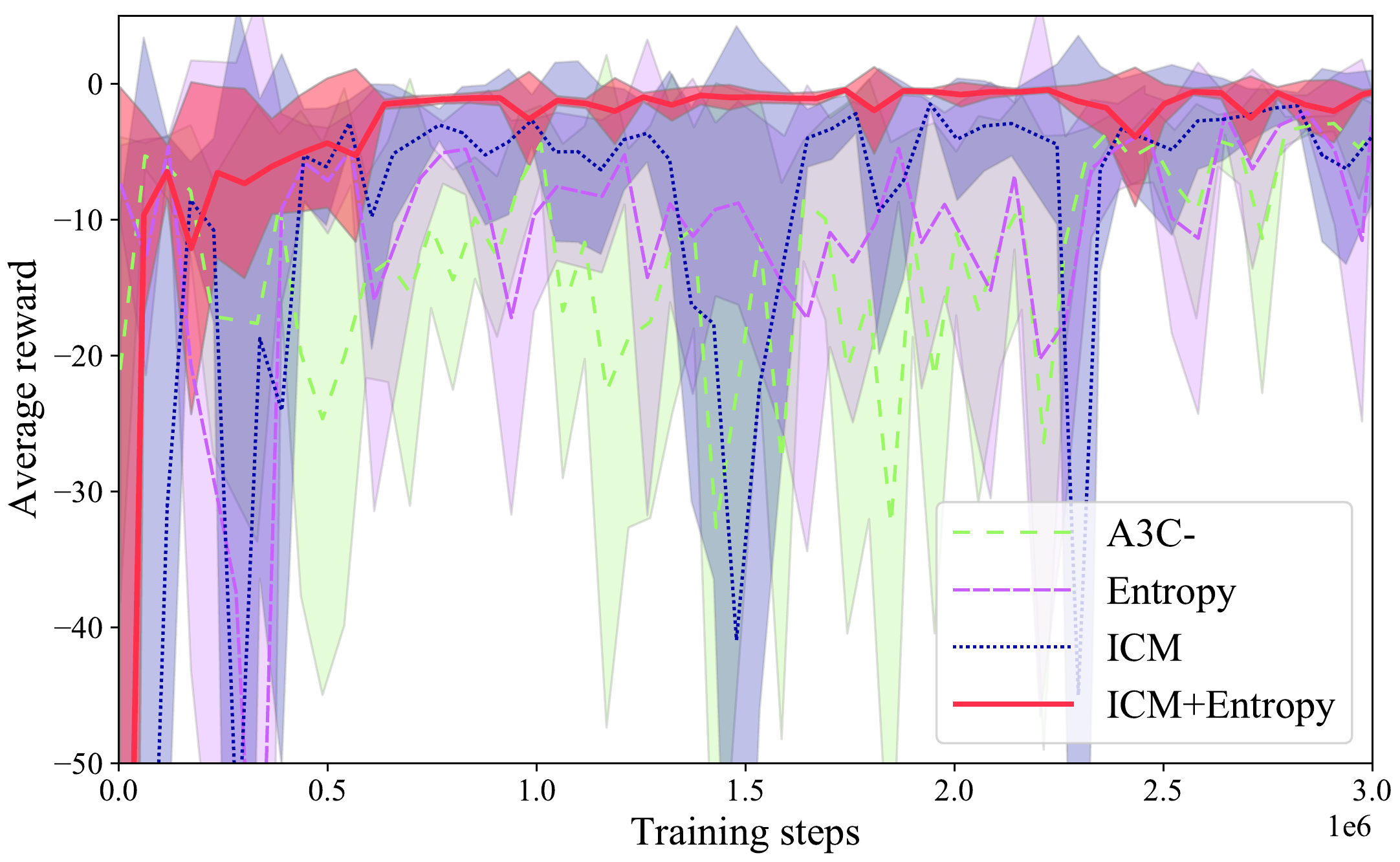}
            \caption{Average reward.}
            \label{fig:reward}
        \end{subfigure}%

        \begin{subfigure}[t]{0.9\linewidth}
            \centering
            \includegraphics[width=\linewidth]{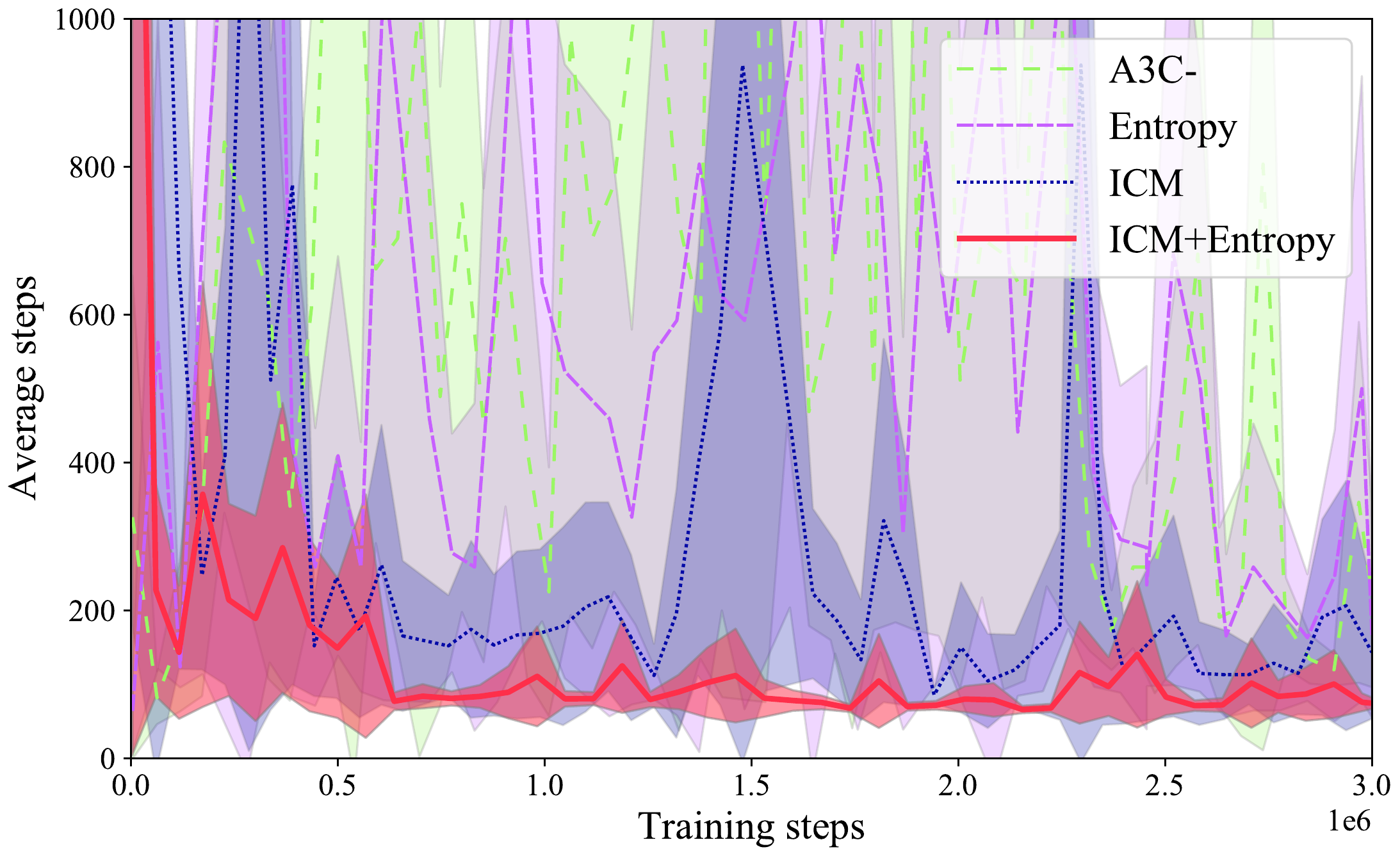}
            \caption{Average steps.}
            \label{fig:steps}
        \end{subfigure}
    \caption{Comparison of (Fig. \ref{fig:reward}) average reward and (Fig. \ref{fig:steps}) steps 
            for different exploration strategies that we considered
            (switching on/off the entropy loss or the intrinsic reward).
            The mean and confidence interval for each configuration
            is calculated over the statistics of 3 independent runs.
            }
    \label{fig:visplots}
\end{figure}

The state $\bs_t$ consists of $72$-dimensional laser range readings $\bs^l_t$
(with a maximum range of $\SI{7}{\metre}$),
and a $3$-dimensional relative goal position $\bs^g_t$
(relative distance, and $\sin$ and $\cos$ of the relative orientation in the agent's local coordinate frame).
At each timestep, the agent can select one of three discrete actions:
\{go straight for \mbox{$\SI{0.06}{\metre}$}, turn left $8$\textdegree, turn right $8$\textdegree\}.

The hyper-parameters regarding the reward function (\ref{sssec:external},\ref{sssec:intrisic}) are the following:
$\gamma=0.99$,
$\lambda^p = 1$,
$\lambda^{\omega} = \frac{1}{200\pi}$,
$\lambda^g = 0.15$,
$r_{\text{reach}} = 1$,
$r_{\text{collision}} = -5$,
$r_{\text{position}} = -0.05$,
and the scale for the intrinsic reward $\lambda^i$ is set to $1$.

The actor-critic network uses two convolutional layers of $8$ filters with stride $2$,
and kernel sizes of $5$ and $3$ respectively,
each followed by ELU nonlinearities.
These are followed by two fully connected layers with $64$ and $16$ units,
also ELU nonlinearities,
to transform the laser sensor readings $\bs^l_t$ into a $16$-dimensional embedding.
This representation is concatenated with the relative goal position $\bs^g_t$,
and fed into a single layer of $16$ LSTM cells.
The output is then concatenated with $\bs^g_t$ again,
and used to produce the discrete action probabilities
by a linear layer followed by a softmax,
and the value function by a linear layer.

The ICM inverse model $\psi^i$ consists of three fully connected layers
with 128, 64, and 16 units,
each followed by an ELU nonlinearity,
producing features $\phi_t$ and $\phi_{t+1}$ from $\bs^l_t$ and $\bs^l_{t+1}$.
Those features are then concatenated and put through a fully connected layer with 32 units and an ELU nonlinearity,
the output of which predicts $\hat{\ba}_t$ by a linear layer followed by a softmax.
The ICM forward model $\psi^f$ accepts the true features $\phi_t$ and a one-hot representation of the action $\ba_t$,
and feeds them into two linear layers with 64 and 32 units respectively, followed by ELU
and a linear layer \mbox{that predicts $\hat{\phi}_{t+1}$}.

We train A3C with the Adam optimizer with statistics shared across $22$ learner threads,
with a learning rate of $1\mathrm{e}{-4}$.
The rollout step $K$ is set to $50$.
ICM is learned jointly with shared Adam with the same learning rate and $\lambda^f = 0.2$.
Each training of $3$ million iterations takes approximately $15$ hours using only CPU computation.

We only train agents on \textit{Map1} (Fig. \ref{fig:map1}),
varying exploration strategies (Sec. \ref{sec:map1}).
\textit{Map2,3,4} are used to test the generalization ability of the trained policies (Sec. \ref{sec:map234}).

\subsection{Training and Evaluation on \textit{Map1}}
\label{sec:map1}

We experiment with $4$ variations of exploration strategies by switching on and off the entropy loss and the intrinsic reward:
1) \textit{A3C-}: A3C with $\beta=0.$;
2) \textit{Entropy}: A3C with $\beta=0.01$;
3) \textit{ICM}: A3C with $\beta=0.$, with ICM;
\mbox{4) \textit{ICM+Entropy}}: A3C with $\beta=0.01$, with ICM.

The average reward and steps obtained in the evaluation during training are shown in Fig. \ref{fig:visplots}.
We can clearly observe that without any explicit exploration mechanisms,
\textit{A3C-} exhibits unstable behavior during learning.
\textit{ICM} alone is much more effective in guiding the policy learning than \textit{Entropy} alone,
since the exploration imposed by the latter is state-independent,
while with \textit{ICM}
the agent is able to actively acquire new useful knowledge from the environment,
and learn efficient navigation strategies for the environmental structures that it has encountered.
\textit{ICM+Entropy} stably and efficiently achieves the best performance.

We run evaluations on the same set of $300$ random episodes on \textit{Map1} after training,
and report the resulting statistics in Table \ref{tab:evaluation}.
In addition,
we conducted another set of experiments
where we remove the LSTM layers in the actor-critic network.
We report their evaluation results in Table \ref{tab:evaluation} as well.
Although without LSTM
none of the configurations manage to converge to a good policy
(so we omit their training curves in Fig. \ref{fig:visplots}),
we can still observe a clear trend of performance improvements
for the policies trained both with and without LSTM,
brought about by ICM exploration.

\begin{table}[!t]
\caption{Evaluation on \textit{Map1}.
}
\label{tab:evaluation}
\begin{center}
\begin{tabular}{c|ccc}
\hline
\hline
                                &    Exploration    &  Success             & Steps  \\
                                &    Strategy       &  Ratio (\%)          & (mean$\pm$std)  \\
\hline
 \multirow{2}{*}{\textit{Map1}} &    Entropy        & 34.3                 & 313.787$\pm$135.513 \\
                                &    ICM            & 65.7                 & 233.147$\pm$143.260 \\
           (w/o LSTM)           &    ICM+Entropy    & \textbf{73.3}        & \textbf{162.733$\pm$150.291} \\
\hline
 \multirow{2}{*}{\textit{Map1}} &    A3C-           & 88.3                 & 173.063$\pm$123.277  \\
 								&    Entropy        & 96.7                 & 102.220$\pm$90.230  \\
                                &    ICM            & 98.7                 & 91.230$\pm$62.511  \\
          (w/ LSTM)             &    ICM+Entropy    & \textbf{100}         & \textbf{75.160$\pm$52.075} \\
\hline
\end{tabular}
\end{center}
\end{table}

\subsection{Generalization Tests on \textit{Map2,3,4}}
\label{sec:map234}

To test the generalization capabilities of the learned policies,
we deploy the networks trained on \textit{Map1}, collect statistics on a fixed set of $300$ random episodes
on \textit{Map2,3,4},
and report the results in Table \ref{tab:test}.

As discussed earlier in Fig. \ref{fig:maps},
\textit{Map2} is relatively simple but contains unfamiliar structures;
\textit{Map3} and \textit{Map4} are more challenging,
but the structures inside these floorplans are more similar to the training environment of \textit{Map1}.
With the exploratory behaviors brought by both \textit{ICM} and \textit{Entropy},
the agent learns well how to exploit structures similar to those it has encountered during training.
Thus from Table \ref{tab:test} we can observe that,
although the full model \textit{ICM+Entropy} is slightly worse
than \textit{ICM} exploration alone on \textit{Map2},
it achieves better performance in the more challenging environments of \textit{Map3} and \textit{Map4}.

\begin{table}[t]
\caption{Generalization tests on new maps.}
\label{tab:test}
\begin{center}

\begin{tabular}{c|ccc}
\hline
\hline
                                &    Exploration             &  Success             & Steps                        \\
                                &    Strategy                &  Ratio (\%)          & (mean$\pm$std)               \\
\hline
\multirow{3}{*}{\textit{Map2}}  &    A3C-                    &  66.0                & 199.343$\pm$145.924          \\
								&    Entropy                 &  85.3                & 115.567$\pm$107.474          \\
                                &    ICM                     &  \textbf{87.7}       & \textbf{101.160$\pm$102.789} \\
                                &    ICM+Entropy             &  82.7                & 109.690$\pm$107.043          \\
\hline
\multirow{3}{*}{\textit{Map3}}  &    A3C-                    &  51.3                & 197.200$\pm$141.292          \\
								&    Entropy                 &  69.7                & 175.667$\pm$139.811          \\
                                &    ICM                     &  63.7                & 152.627$\pm$135.982          \\
                                &    ICM+Entropy             &  \textbf{72.7}       & \textbf{139.423$\pm$120.144} \\
\hline
\multirow{3}{*}{\textit{Map4}}  &    A3C-                    &  44.3                & 273.353$\pm$148.527          \\
								&    Entropy                 &  34.3                & 240.197$\pm$164.565          \\
                                &    ICM                     &  45.0                & 236.847$\pm$169.169          \\
                                &    ICM+Entropy             &  \textbf{55.0}       & \textbf{209.500$\pm$161.767} \\
\hline
\end{tabular}
\end{center}
\end{table}

\subsection{Discussion}
\label{sec:disc}

As can be seen from our results, the ICM agents, even if they are not always more successful in reaching the targets, tend to find shorter paths to them. We believe that this can be viewed as evidence that exploratory behaviors due to curiosity help escape local minima of the policy function (in form of solutions that choose inefficient ways of reaching the target). We speculate that the reason for this improvement lies in the ICM agents seeing a bigger part of the environment, as the intrinsic motivation attracts them towards novel and currently less predictable states, while random exploration does little to avoid re-exploring the same areas again and again, getting stuck in structures which are difficult to escape by pure chance.


\section{Conclusions}
\label{sec:conclusions}

We investigate exploration strategies for learning mapless navigation policies with DRL,
augmenting the reward signals with intrinsic motivation.
Our proposed method achieves better sample efficiency and convergence properties during training than all the baseline methods,
as well as better generalization capabilities during tests on previously unseen maps.
Our future work includes studying how to better adapt the weighting of the different components in the reward function,
as well as real-world robotics experiments.

\addtolength{\textheight}{-12cm}   



%
%
%

\bibliographystyle{IEEEtran}
\bibliography{zheloo18icraws}

\end{document}